\documentclass{article}

\usepackage{arxiv}
\usepackage{cite}
\usepackage{amsmath,amssymb,amsfonts}
\usepackage{algorithmic}
\usepackage{textcomp}
\usepackage[english]{babel}
\usepackage{latexsym}
\usepackage{url}

\usepackage[utf8]{inputenc} % allow utf-8 input
\usepackage[T1]{fontenc}    % use 8-bit T1 fonts
\usepackage{hyperref}       % hyperlinks
\usepackage{url}            % simple URL typesetting
\usepackage{booktabs}       % professional-quality tables
\usepackage{amsfonts}       % blackboard math symbols
\usepackage{nicefrac}       % compact symbols for 1/2, etc.
\usepackage{microtype}      % microtypography
\usepackage{lipsum}
\usepackage{graphicx}
\graphicspath{ {./images/} }

\title{ConvTransSeg: A Multi-resolution Convolution-Transformer Network for Medical Image Segmentation}

\author{
 Zhendi Gong \\
  School of Computer Science\\
  University of Nottingham\\
  Nottingham, NG8 1BB \\
  \texttt{psxzg6@nottingham.ac.uk} \\
   \And
 Andrew P. French \\
  School of Computer Science\\
  University of Nottingham\\
  Nottingham, NG8 1BB \\
  \And
 Guoping Qiu \\
  School of Computer Science\\
  University of Nottingham\\
  Nottingham, NG8 1BB \\
  \And
 Xin Chen \\
  School of Computer Science\\
  University of Nottingham\\
  Nottingham, NG8 1BB \\
  \texttt{xin.chen@nottingham.ac.uk}
}

\begin{document}
\maketitle
\begin{abstract}
Convolutional neural networks (CNNs) achieved the state-of-the-art performance in medical image segmentation due to their ability to extract highly complex feature representations. However, it is argued in recent studies that traditional CNNs lack the intelligence to capture long-term dependencies of different image regions. Following the success of applying Transformer models on natural language processing tasks, the medical image segmentation field has also witnessed growing interest in utilizing Transformers, due to their ability to capture long-range contextual information. However, unlike CNNs, Transformers lack the ability to learn local feature representations. Thus, to fully utilize the advantages of both CNNs and Transformers, we propose a hybrid encoder-decoder segmentation model (ConvTransSeg). It consists of a multi-layer CNN as the encoder for feature learning and the corresponding multi-level Transformer as the decoder for segmentation prediction. The encoder and decoder are interconnected in a multi-resolution manner. We compared our method with many other state-of-the-art hybrid CNN and Transformer segmentation models on binary and multiple class image segmentation tasks using several public medical image datasets, including skin lesion, polyp, cell and brain tissue. The experimental results show that our method achieves overall the best performance in terms of Dice coefficient and average symmetric surface distance measures with low model complexity and memory consumption. In contrast to most Transformer-based methods that we compared, our method does not require the use of pre-trained models to achieve similar or better performance. The code is freely available for research purposes on Github: (the link will be added upon acceptance). 
\end{abstract}


\twocolumn
\section{Introduction}
\label{sec:introduction}
Accurate image segmentation is essential for many clinical applications, such as computer-aided diagnosis, treatment planning and image-guided surgery \cite{shamshad2022transformers}. Conventional methods require detailed domain knowledge from human and task specific designs, such as active contours \cite{kass1988snakes}, atlas-based methods \cite{gee1993elastically} and graph cuts \cite{boykov2000interactive}. These methods do not require a large number of training examples. However, due to complex parameter settings and poor robustness on different imaging conditions (e.g intensity variations, presence of noise, etc.), these methods were outperformed by deep-learning-based methods, particularly convolutional neural networks (CNNs) \cite{zhou2021review}.

In CNNs, convolutional filters are applied to local image regions in a hierarchical manner, which makes them effective in learning multi-resolution image features and leads to their success in many computer vision tasks (e.g. object recognition and segmentation). With the growth of computing ability, CNNs have become the backbone for most medical image analysis tasks, including image segmentation \cite{ramachandran2019stand}. One of the most widely used CNNs in medical image segmentation is U-Net \cite{ronneberger2015u}, due to its effective learning capability from a smaller number of training examples compared to other CNN models (e.g., Fully Convolutional Network \cite{long2015fully}). U-Net consists of an encoder to extract multi-resolution image features, a decoder to up-sample the feature maps for segmentation prediction and skip connections between the encoder and decoder for a more efficient model learning. 

Many efforts have been made to improve the vanilla U-Net. The U-Net variants include: Residual U-Net (ResUNet) \cite{zhang2018road}, which applied the residual network mechanism \cite{he2016deep} to the U-net for handling vanishing gradients problem; U-Net++ \cite{zhou2018unet++}, which applied more complex skip connections to utilize multi-scale information better; 3D U-Net \cite{cciccek20163d} and V-Net \cite{milletari2016v}, which extended the U-Net to 3D segmentation tasks, such as MRI volumes segmentation. Although the image segmentation field witnessed the success of CNNs, such methods still suffer from the weak ability of capturing long-range regional interactions, which is important in certain applications. With the lack of global contextual information of the whole image, CNNs may produce sub-optimal results (shown in section \ref{sec:Experiments and results}). 

More recently, inspired by the work of Vaswani et al. \cite{vaswani2017attention}, which aims to capture long-range context in machine translation tasks, Vision Transformer (ViT) was proposed by Dosovitskiy et al. \cite{dosovitskiy2020image} to perform image classification and has achieved the state-of-the-art performance in many vision tasks. It treats the whole image as a sentence and treats the divided image patches as different words in a sentence. Specifically, they applied several encoders from Transformer as attention layers to process the linearly transformed inputs. The results have shown that ViT can alleviate the inductive bias of CNNs \cite{dosovitskiy2020image}. Due to the success of ViT in image classification tasks, many efforts have been made to apply the attention layers from ViT in medical image segmentation tasks \cite{chen2021transunet,valanarasu2021medical,wu2022fat}. However, most of these methods require pretraining due to the poor ability of Transformers to learn from small dataset.

Inspired by CNNs and ViT, our proposed method utilizes the advantages of both methods. Due to the effective feature learning property provided by CNN, our method applies CNN to extract multi-resolution image features, which is considered as an encoder. These learned features represent a rich characteristic of local image patches, which are then input to a set of attention blocks in a Transformer, which forms the decoder of the proposed segmentation model. Both the encoder and decoder are multi-resolution which are interconnected seamlessly via computationally efficient re-sampling and re-shaping operations. This novel architecture design enables both effective local feature learning and long-range contextual information interaction for image segmentation. For method comparison, we evaluated our method and many other state-of-the-art methods on several publicly available datasets for binary-class segmentation, including skin lesion dataset (ISIC2018 \cite{codella2019skin}), cell dataset (Bowl: \url{https://www.kaggle.com/c/data-science-bowl-2018}), polyp dataset (CVC-ClinicDB \cite{bernal2015wm}), and multi-class segmentation, including cell dataset (Pannuke \cite{gamper2019pannuke}) and brain tissue dataset (OASIS-1 \cite{marcus2007open}).

\section{Related works}
\label{sec:Related works}
Transformer was initially proposed in 2017 by Vaswani et al. \cite{vaswani2017attention} to perform the machine translation task in natural language processing. The basic building block in a Transformer is self-attention, which is used to capture long-term dependencies between the input tokens (e.g. words). Specifically, self-attention aims to calculate the weighted sum of the values, in which the weight for each value is computed by a function of a query-key pair, where values (V), queries (Q) and keys (K)  are all input vectors. For instance, when the self-attention is applied in machine translation, V, Q and K are the embedding vectors for the input words. Within the self-attention layer, all positions are connected with a number of operations \cite{vaswani2017attention}, whereas a convolutional layer in CNN does not connect all paired positions between the input and output. Additionally, they used a multi-head attention (MHA) block instead of a single attention function based on the finding that it is beneficial to firstly project V, Q and K by different learnable linear layers several times in parallel attention functions. The model is able to manage information from various representation subspaces concurrently with multi-head attention. As the attention block can be taught to focus on specific regions within a context, it becomes a crucial part of neural transduction models for many techniques \cite{chorowski2015attention,devlin2018bert,wu2016google}.

Based on the architecture of encoders in Transformer, Dosovitskiy et al. \cite{dosovitskiy2020image} proposed a Vision Transformer (ViT) for image classification tasks. Different from other works applying attention mechanism in vision tasks, the image-specific inductive bias was not introduced in their work, they directly used a standard Transformer encoder from the work of Vaswani et al. \cite{vaswani2017attention}. They firstly divided the input image into several contiguous patches with the same size. Additionally, to apply the same attention layers in Transformer, they flattened the 2D image patches into 1D vectors as the input tokens. Meanwhile, they projected the vectors to a subspace with learnable linear layers. In other words, V, Q and K are represented by the embedding vectors of image patches in their method. However, the success of ViT is highly dependent on pre-trained models and large training datasets. The limited availability of medical data restricts the application of ViT in the field of medical image processing. After Transformer and ViT were proposed, the combination of CNN and attention mechanisms to achieve medical image segmentation has been an active research area. We divide the recently proposed methods into four categories, dependent on the relative locations of the Transformer and CNN: \textit{Transformer in encoder},  \textit{Transformer as a bridge}, \textit{ Transformer in decoder}, \textit{Fused CNN and Transformer}. Besides, some methods also applied pure Transformer models to perform medical image segmentation; we also presented some of these methods.

\subsection{Transformer in encoder}
Hatamizadeh et al. \cite{hatamizadeh2022unetr} proposed UNETR for 3D medical image segmentation. They firstly divided the 3D volume into 3D image patches and input them into the Transformer encoder.  The multi-scale outputs were further processed by CNN encoder-decoder architecture, where skip connections were applied at different feature scales. Another work is from Wu et al. \cite{wu2022fat}, who proposed a feature adaptive transformer, named  FAT-net, which applied a dual encoder by using both CNN and Transformer to perform feature learning. Additionally, they applied a feature adaptive module in the skip connection to perform feature fusion and a memory-efficient decoder to further process the output feature maps for segmentation prediction. Additionally, Li et al. \cite{DBLP:conf/miccai/QiY0LWL019} proposed X-Net, which contains a CNN branch and a Transformer branch. The CNN branch takes the original image as input and performs reconstruction, and the Transformer branch firstly encodes the divided image patches and then applies convolutional layers from the CNN branch to decode the output. Shen at al \cite{shen2022automated} proposed a model called COTR-Net to perform automated kidney tumor segmentation, where Transformer layers are inserted after each CNN block in the encoder of a U-Net-like model.

\subsection{Transformer as a  bridge}
TransUNet \cite{chen2021transunet} used an attention module consisted of 12 Transformer layers to capture the long-term relationship of the hidden features extracted by a CNN. The decoder applied CNN-based up-sampling layers to process the output from the attention module with skip connections applied between the encoder and decoder. TransAttUNet \cite{chen2021transattunet} applied both Transformer Self Attention and Global Spatial Attention between the encoder and decoder. Additionally, they applied multi-scale skip connections to enhance the performance of the original U-shaped architecture. Similarly, Ji at al. \cite{ji2021multi} proposed a Multi-Compound Transformer which applied two separate Transformer modules between the encoder and decoder. Specifically, they used a Transformer self-attention module to construct the multi-scale context and applied a proxy-embedding in the Transformer cross-attention module to handle feature representations and category dependencies. Additionally, Zhang at al. \cite{zhang2021multi} applied Transformer blocks as a bridge between the CNN-based encoder and decoder to perform cell segmentation. After the decoder, they combined different losses on the cell body and edge to achieve better performance on learning separate information, which can provide accurate information of the edges and support local consistency. Different from the above methods that apply Transformer blocks as a bridge to connect the encoder and decoder, AttUnet \cite{oktay2018attention} applied attention gates (AG) to filter the encoded features through the skip connections. They aimed to reduce the false-positive predictions for small objects using the feature selectivity offered by AG.

\subsection{Transformer in decoder}
\label{sec:Transformer in decoder}
Different from the aforementioned methods, Transformers can also be applied after the CNN encoder as a decoder to capture the global context information of the CNN encoded features. However, there are very limited works in this category. Li at al. \cite{li2021medical} proposed SegTran for medical image segmentation. They applied a pre-trained ResNet in the encoder to extract the feature maps and used the squeeze-and-expansion Transformer in the decoder, with a squeeze block to learn attention matrix and an expansion block to learn varied representations. Another work is from Li et al. \cite{li2021more}, who proposed a window attention up-sampling module in the classic U-Net decoder to replace the original convolutional up-sampling method. Specifically, the different resolution levels in the decoder are connected by the window attention decoder and bilinear up-sampling, which is used as a residual connection.

\subsection{Fused CNN and Transformer}
Li et al. \cite{li2021gt} proposed a U-Net like group Transformer network GT U-Net. It inherits the architecture of the traditional U-Net \cite{ronneberger2015u}, but grouped convolutions were applied instead of the original convolution layers. Additionally, they applied self-attention based Transformer blocks between each pair of group convolutions to learn the global dependencies of different feature representations. In another work, named as TransFuse \cite{zhang2021transfuse}, a convolutional branch and a Transformer branch are fused parallelly by their proposed BiFusion module. Specifically, the Transformer branch takes image patches as input then reshapes and up-samples the output to different scales. Meanwhile, the output features with different resolution from a CNN branch, which takes the whole image as input, are fused with the outputs from a Transformer branch. Furthermore, Gao, et al. proposed UTNet \cite{gao2021utnet}, which applies Transformer blocks after the convolutional blocks both in CNN encoder and decoder. Additionally, skip connections are used between encoder and decoder at each resolution level.
\begin{figure*}[!]
\centering
\includegraphics[width=0.8\linewidth]{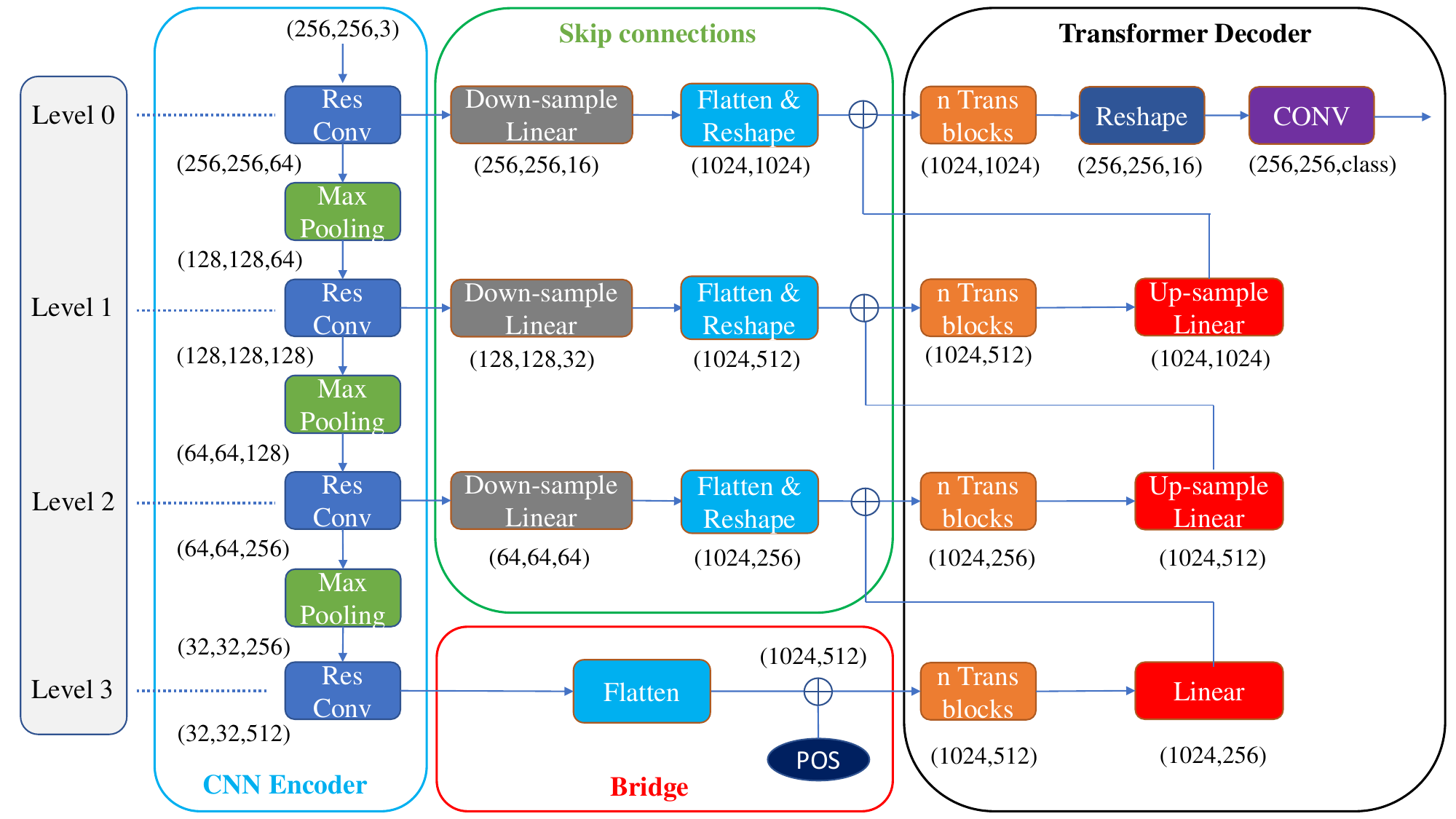}
\caption{\label{fig1}The overall model structure of the proposed ConvTransSeg for a multi-class image segmentation task. The example uses an input of 3-channel $256\times256$ color Image. The number of model levels is 4 and the number of feature channels at level 0 is 64.}
\end{figure*}

\subsection{Pure Transformer}
Karimi et al. \cite{karimi2021convolution} proposed a convolution-free method for 3D medical image segmentation. They flattened the divided 3D image patches from the original volume to 1D vectors and input them to the Transformer mechanism. After a set of attention functions, they pick the output of the center patch and reshape it as the segmentation prediction on that patch. Furthermore, inspired by the U-Net architecture, Cao et al \cite{cao2021swin} proposed Swin-Unet to perform medical image segmentation using the Swin Transformer \cite{liu2021swin} as the basic block. Different from the regular Transformer block used in ViT, the Swin Transformer uses shifted windows to perform self-attention computation, which has a higher computational efficiency. Swin-Unet applied these Swin Transformer in a U-shape architecture with skip connections between the encoder and decoder. MISSFormer \cite{huang2021missformer} has a similar architecture with Swin-Unet with Swin Transformer and regular skip connections replaced by an enhanced Transformer block and an enhanced Transformer context bridge.

\subsection{Contributions}
It can be concluded from the above related works that the mainstream idea is to use a Transformer’s attention mechanism to capture the long-range relationships of image patches either using their raw values or the feature representations derived from CNNs. The main novelties and contributions of our work are summarized as below:
\begin{enumerate}
\item Our proposed image segmentation method is a hybrid model which applies convolutions as local image feature extractors in an encoder and attention layers in a decoder for segmentation prediction. Compared to the works \cite{li2021medical,li2021more} that also apply the Transformer as decoder, our method has certain advantages. Specifically, Li, et al \cite{li2021more} only used attention layers to replace the original up-sampling layers in the decoder of U-Net. Our method explores the feasibility of utilizing a pure Transformer in the entire decoder, since we aim to fully use the features of Transformer to capture the long-term dependencies between the local features of different image patches. Additionally, although SegTran \cite{li2021medical} used a pure Transformer in the decoder, their  Transformer layers only used the final output of a CNN encoder without multi-resolution information. We propose a \textit{new approach to connect a CNN-based encoder and a pure Transformer-based decoder in multi-resolution} that are seamlessly connected in their corresponding levels, including dimensionality reduction and reshaping operations. No such efficient connection between CNN encoder and Transformer decoder has been proposed in the literature.
\item Many Transformer-based methods require pretraining, since Transformers lack the ability to work well on limited datasets. However, our method \textit{does not require any pretraining}, even when training on a small dataset, and can still achieve similar or better performance compared to the models with pretraining. Without the need for pretraining, the architecture of our model allows us to use different image sizes as the input as an additional bonus.
\end{enumerate}

\section{Method}
\subsection{Overall model structure}
The overall model structure of our method is depicted in Fig \ref{fig1}. Our method contains four main components, including a CNN-based encoder (CE), a Transformer-based decoder (TD), and skip connections and a bridge between CE and TD. The CE follows a similar design of the encoder in ResUNet \cite{zhang2018road}, where each convolutional block is a residual convolutional block (ResConv). A max pooling layer is applied after each ResConv block to down-sample the learned feature maps into lower image resolutions. The TD is also multi-resolution. At each level, $n$ cascaded self-attention blocks (Trans blocks) are designed to take the transformed CNN-features from the corresponding level as the input. Additionally, a linear layer (Linear or Up-sample Linear) is applied to project the features of the current level to the same vector dimension of the higher resolution level, which is then added to the higher dimensional feature vector as the input to TD. Moreover, skip connections that contain linear down-sampling (Down-sample Linear) and reshaping (Flatten \& Reshape) operators, and a bridge that contains flattening (Flatten) and a positional embedding (POS) are applied to connect the CE and TD in the corresponding levels. Finally, a reshaping (Reshape) and convolutional layer (CONV) in TD are applied to convert the output of the final Transformer blocks to multi-class segmentation predictions.

Although the input image size is flexible in our method, to assist the understanding of the model structure, an input color image (3 RGB channels) with size of 256×256 is used as an example to illustrate the dimensions of data at each of the building blocks in a 4-level architecture as shown in Fig \ref{fig1}. The justifications and working principles of the proposed design are provided in the following subsections.  

\subsection{CNN encoder}
\begin{figure}[!t]
\centering
\includegraphics[width=0.7\linewidth]{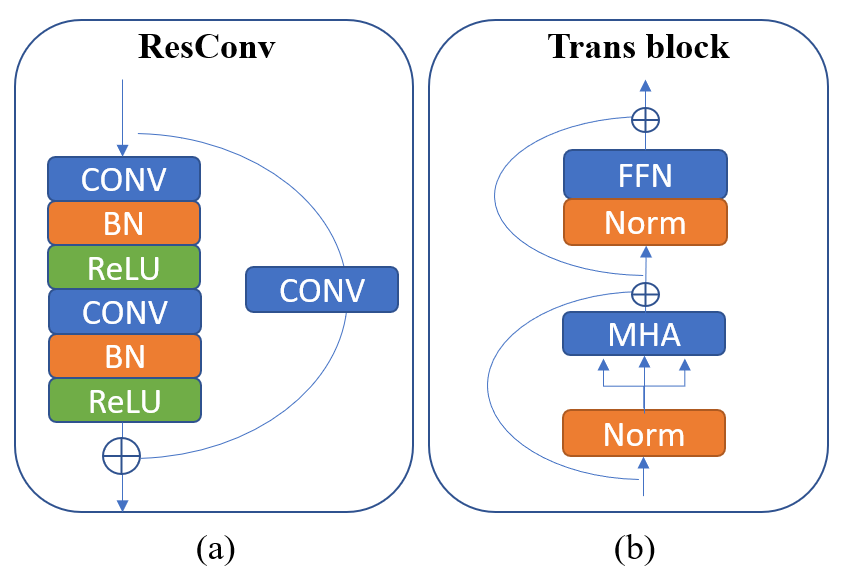}
\caption{\label{fig2}The key components in (a) the ResConv block, (b) the Trans block.}
\end{figure}

A residual block based CNN\cite{zhang2018road} is adopted as the encoder for feature extraction in our method (Fig \ref{fig1}: CNN Encoder- CE). As shown in Fig \ref{fig2} (a), each ResConv block consists of two convolutional layers with the kernel size of 3×3 using zero padding, each followed by a batch normalization layer and a ReLU activation layer. The first convolutional layer in each ResConv block doubles the number of feature channels of the input data, except for the block in the first level, where the first convolutional layer changes the number of feature channels of the input data to  $C_{base}$ (e.g. $C_{base}=64$ in Level 0 in Fig \ref{fig1}). The second convolutional layer in each ResConv block has the same number of feature channels as the first convolutional layer. Additionally, a single convolutional layer is added as the skip connection between the output and the input of ResConv. Specifically in the CNN Encoder, the input image ${X_{in}\ \in R}^{W\times H\times C_{ini}}$ (\textit{W}, \textit{H}, and \textit{$C_{ini}$} indicate the width, height, and the number of channels respectively of the input image) is processed to generate ${X_{out_i}^{CE}\ \in\mathbb{R}}^{(W/2^i)\times(H/2^i)\times{(2^iC}_{base})}$ in each level \textit{i} (\textit{i} = 0, 1, 2, …, \textit{l}, \textit{l} indicates the depth of CE)

\subsection{Skip connections and bridge}
The output of the deepest level in CE ${X_{out_l}^{CE}\ \in\mathbb{R}}^{(W/2^l)\times(H/2^l)\times({2^lC}_{base})}$ is flattened spatially to $\mathbb{R}^{P_l\times(2^lC_{base})}$ and added to a learnable positional embedding that has the same dimension, where $P_l=(W/2^l)\ast(H/2^l)$. This process aims to learn the positional information of the input feature vectors (tokens), before being input to the TD. In other words, the input to TD in level $l$ can be considered as the feature embeddings of $P_l$ image patches, each with the size of  $2^l\times2^l$. 

Instead of only using the lowest feature resolution at level $l$, the skip connections between CE and TD aim to help the Transformer layers learn the CNN features from different resolution levels. However, if the output of CE is flattened and input to TD directly in all levels, the whole model would be extremely large to be computed (e.g. it would require W×H tokens in level 0). To solve such a problem, we fix the number of input tokens to $P_l$ in the TD for all levels. Thus, the TD can be considered as always processing the feature vectors of $2^l\times2^l$ patches of the input image.  Subsequently, a learnable linear layer is added (the "Down-sample linear" shown in Fig \ref{fig1}) to down-sample the feature channels of the CNN outputs before reshaping them to the desired feature dimensions. Specifically, in level $i$ ($i<l$), the output from CE ${X_{out_i}^{CE}\ \in\mathbb{R}}^{(W/2^i)\times(H/2^i)\times{(2^iC}_{base})}$ is first linearly down-sampled to $\mathbb{R}^{(W/2^i)\times(H/2^i)\times{(2^iC}_{base}/m)}$, where $m$ indicates the down-sampling factor that is consistent for all levels ($m$=4 in the example shown in Fig \ref{fig1}). Subsequently, this three-dimensional tensor at each level is reshaped to $P_l$ (e.g. $P_l=1024$ in Fig \ref{fig1}) by $d_i$ (e.g. $d_0=1024$, $d_1=512$, $d_2=256$ in Fig \ref{fig1}). This “Flatten \& Reshape” process effectively rearranges the learned CNN features at each level to $P_l$  image patches with a feature vector length of $d_i$. 

As the example shown in Fig \ref{fig1}, the feature map from CE has the dimension of $64\times64\times256$ in level 2. The down-sampling linear layer in level 2 is then firstly reduce the dimension of the feature channels from 256 to 64 for dimensionality reduction purpose. The first two dimensions $64\times64$ can be considered as $32\times32$ (1024) image patches, each with a size of $2\times2$. The “Flatten \& Reshape” process then coverts this $64\times64\times64$ 3D feature to a 2D $1024\times256$ feature representation. 1024 is the number of image patches, and each has the length of $64\times2\times2$ (concatenating the feature vectors in a $2\times2$ image region). As the resolution increases, with a fixed number of image patches (1024), the sizes of image patches increase hence $d_i$ also increases. The output of the “Flatten \& Reshape” process in level $i$ is then added to the output feature vectors from TD in level $i+1$ to form the input to TD in level $i$. 

\subsection{Transformer decoder}
\begin{figure}[!t]
\centering
\includegraphics[width=0.6\linewidth]{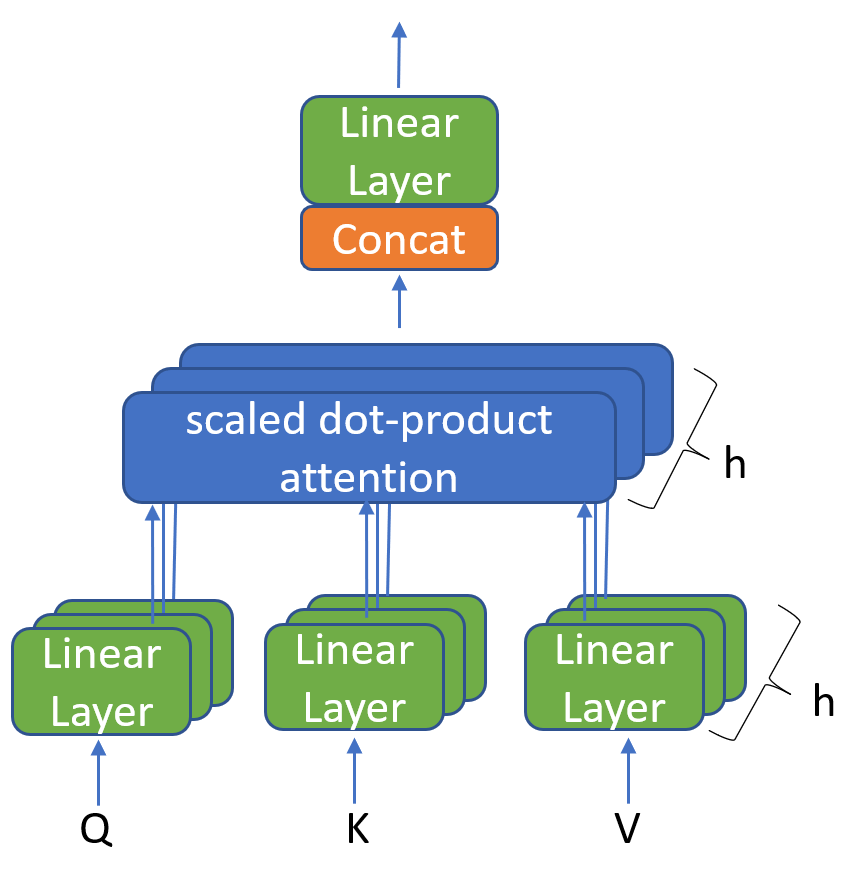}
\caption{\label{fig3}The key components of the multi-head attention (MHA) block.}
\end{figure}

After feature extraction in the CE and feature remapping in the skip connections, TD is responsible for decision making in predicting segmentation output. Each level in TD contains $n$ Transformer blocks to process the output features from the bridge and skip connections, and is followed by a linear layer to project the feature vector from the current level $i$ to the $i-1$ level. The Transformer blocks make no changes to the dimension of the feature vector, while the linear layers change the feature vectors’ dimension to the dimension of the TD input in lower level. The Transformer block is shown in Fig \ref{fig2} (b), which was originally introduced by Vaswani et al. \cite{vaswani2017attention}. The block contains a multi-head self-attention mechanism (MHA) and a position-wise fully connected feed-forward network (FFN). Additionally, a normalization layer is added before MHA and FFN, and a dropout layer is added after MHA and FFN. Furthermore, a skip connection is used between the input and the dropout layer. Thus, the calculation within a Transformer block is expressed as below:
\begin{equation}
\begin{aligned}
X_{out1}=X_{TransIn}+Dropout(MHA(Norm(X_{TransIn}),\\
Norm(X_{TransIn}),Norm(X_{TransIn})))
\end{aligned}
\end{equation}
\begin{equation}
X_{out}=X_{out1}+Dropout(FFN(Norm(X_{out1})))
\end{equation}
where $X_{trans_in}$ and $X_{out}$ are the input and output of the block. Instead of performing a single attention mechanism, the MHA calculates the scaled dot-product attention (SDPA) several times in parallel, as illustrated in Fig \ref{fig3}. It takes $Q$, $K$ and $V$ as input, and outputs the weighted sum of $V$. Since we use self-attention in the whole method, the matrices $Q$, $K$, and $V$ are the input feature vectors to the TD, which are identical to each other and are consistently updated during model training. As illustrated in Fig \ref{fig3}, they are linearly mapped to $\mathbb{R}^{n_i\times d_i\ }$from $\mathbb{R}^{n_i\times d_i\ }$and reshaped to $\mathbb{R}^{n_i\times d_h\times h\ }$before being input to SDPA. $n_i$ indicates the number of tokens ($P_l$ as described in section \ref{sec:Transformer in decoder}). $d_i$ indicates the token size and h is the number of heads. The new token size $d_h$ is equal to $d_i/h$. In our method, we set $d_h$  to $C_{base}$ and $h$ is calculated as $d_i/d_h$ in all Transformer blocks to keep the number of operations of each head remain the same in different levels. Additionally, all pairs of ${Q_h\ \in\mathbb{R}}^{n_i\times d_h\ }$, ${K_h\ \in\mathbb{R}}^{n_i\times d_h\ }$ and ${V_h\ \in\mathbb{R}}^{n_i\times d_h\ }$are simultaneously processed by SDPA, the outputs  ${V_{out}\ \in\mathbb{R}}^{n_i\times d_h\ }$from each pair of inputs are concatenated together and linearly projected to ${X_{out}^{MHA}\ \in\mathbb{R}}^{n_i\times d_i}$. The calculation in SDPA is shown :
\begin{equation}
W=Softmax\left(\frac{QK^T}{\sqrt{d_i}}\right)
\end{equation}
\begin{equation}
X_{out}^{SDPA}=WV
\end{equation}
where  $d_i$ is the token size of $Q$, $K$ and $V$. Matrix $W$ indicates the weight matrix assigned to $V$, which is used to calculate the weighted sum of $V$. Furthermore, the input of FFN ${X_{in}^{FFN}\ \in\mathbb{R}}^{n_i\times d_i}$ is linearly projected to $\mathbb{R}^{n_i\times{(f\ast d}_i)}$, and processed with a ReLU activation layer and a dropout layer. $f$ is an up-sampling factor which is 2 in our method, so that the FFN can learn information from a higher dimensional space without consuming too much memory. In the last step, the input is projected back to $\mathbb{R}^{n_i\times d_i}$ from $\mathbb{R}^{n_i\times{(f\ast d}_i)}$. In level 0 of the TD, since the CNN output has the same spatial dimension as the original image, instead of using a linear layer, the output ${X_{out}^{TD}\ \in\mathbb{R}}^{P_l\ \times({2^{2l}C}_{base}/m)}$ is firstly reshaped back to $\mathbb{R}^{W\times H\times{(C}_{base}/m)}$ based on the reversed reshaping method in the skip connection . It is then processed by a $1\times1$ convolutional layer to perform predictions, resulting in ${y\ \in\mathbb{R}}^{W\times H\times Class}$ ($Class$ is the number of classes to be segmented). 

Similar to many other works for comparison \cite{chen2021transattunet,ji2021multi,li2021medical,wu2022fat,yan2022after}, the loss function used for training our model is the combination of the cross-entropy loss and the dice loss, expressed as:
\begin{equation}
loss=\ \alpha\ast{loss}_{CE}+\beta\ast{loss}_{Dice}
\end{equation}

\section{Experiments and results}
\label{sec:Experiments and results}
\subsection{Datasets and evaluation metrics}
To evaluate our proposed method and compare it to other state-of-the-art methods, five publicly available medical imaging datasets that cover four medical applications are used:

(1) Skin lesion segmentation (\textbf{ISIC}): ISIC2018 \cite{codella2019skin} (binary-class), which was published by the International Skin Imaging Collaboration challenge, containing 2594 RGB dermoscopy images with different image sizes. 

(2) Cell segmentation: 2018 Data Science Bowl (\textbf{Bowl}): \url{https://www.kaggle.com/c/data-science-bowl-2018} (binary-class) and \textbf{Panuke} \cite{gamper2019pannuke} (5 classes). There are 670 $320\times256$ RGB images in 2018 Data Science Bowl and 7901 $256\times256$ RGB images in Pannuke. The 5 classes for the segmentation mask in Pannuke are Neoplastic (Neo), Inflammatory (Inf), Connective (Con), Dead, and Epithelial (Epi) respectively.

(3) Polyp segmentation (\textbf{CVC}): CVC-ClinicDB \cite{bernal2015wm} (binary-class) contains 612 $384\times288$ RGB colonoscopy images from 31 colonoscopy sequences. 

(4) Brain tissue segmentation (\textbf{OASIS}): OASIS-1 \cite{marcus2007open} (4 classes) contains 414 $192\times192$ gray scale images, the 4 classes for the segmentation mask are Cortex, Subcortical-Gray-Matter (SGM), White-Matter (WM), and Cerebrospinal fluid (CSF) respectively. 

Although our method can accept any size of input image (as long as it can be divided by $2^l$), in our experiments, we reshaped all images to $224\times224$ since most of the methods for comparison used such resolution as the input size (due to they used pre-trained models). Specially, when we compared our model with UTnet \cite{gao2021utnet}, we reshaped the images to $256\times256$, since UTnet used the input size of $256\times256$, and the input size of $224\times224$ cannot fit in the model by their default parameter settings.   The percentage of training, validation and testing sets are 70\%, 10\% and 20\% respectively for all tasks. All models were trained on a 12GB GeForce GTX 1080 Ti GPU with Adam optimizer. Additionally, the learning rate was set to 0.0001 and the models were all trained for 50 epochs. During each training phase, the model with the least validation loss was saved for testing. 

In model evaluation, Dice Coefficient (DC) and Average Symmetric Surface Distance (ASSD) are the main metrics, and we applied Wilcoxon Signed Rank Test (WSRT) to evaluate the statistical significance in difference between two methods. We also compared the number of learnable parameters (NP), the GPU memory usage (GM), and the average training time per step (TS) to evaluate the model complexity and training efficiency. 

It is worth noting that, for the Pannuke dataset, some classes have large number of empty ground truth. Our initial results show that the model tend to consistently generate empty predictions on most images in such classes. Thus, we only calculated the loss on the prediction of the class whose ground truth was not empty in the training process for all methods to avoid model over-fitting caused by unbalanced labels. Similarly, by including these empty classes in evaluation, it results in very high DC scores. It makes the measure being less sensitive to the errors in images with non-empty foreground. Hence, we also only calculated DC and ASSD on non-empty classes (in ground truth mask) in the evaluation.

\subsection{Parameter tuning}
\label{sec:Parameter tuning}
Before comparing our method with other state-of-the-art methods, we tested the sensitivities on the choice of hyper-parameters in our method by performing parameter tuning. Specifically, we set $l=3$ for the depth of CE and TD, so that each feature vector processed in TD is a representation of a $8\times8$ image patch. Additionally, we tested and compared the method performance of using 1, 2, 3, 4, 5 for $n$ (number of Transformer blocks in TD), 32, 64 for $C_{base}$ (number of output feature channels of the first convolutional block) and 2, 4, 8 for $m$ (down-sampling factor). We selected ISIC2018 as a binary-class dataset and OASIS-1 as a multi-class dataset to evaluate the above parameter settings. Considering the trade-off between model performance and model complexity in both datasets, the optimal settings were $n=3$, $C_{base}=64$, and $m=8$. These settings were then consistently used for evaluating all other datasets and compared to other methods.  

\subsection{Ablation study}
The two key novelties of our model design are the integrated multi-level feature fusion and the down-sampling linear layers in skip connections to reduce the model’s complexity. Firstly, to illustrate the benefit of utilizing the output feature maps from CE in multiresolution rather than only using the output of level $l$, we evaluated our method with and without the skip connections (SC) between CE and TD. Secondly, we evaluated our method with and without the down-sampling linear layers (DSL). We choose to use the OASIS-1 dataset in these experiments, since it is a more generic multiclass dataset which includes both large and small objects of interest. We used smaller $C_{base}$ and $m$ values instead of the parameter settings that are concluded in section \ref{sec:Parameter tuning}. Because the model (with SC without DSL) requires significantly larger memory that can not be accommodated by a 12 GB GPU. Thus, we changed $C_{base}$ to 32 instead of 64, which decreases the number of parameters for all models so that the models can be trained on the same GPU. By setting $m$ to 4 instead of 8, it ensures that the model with DSL will not lose much information by the down-sampling process.

\begin{table}[!h]
\caption{\label{tab1}Evaluation of model performance on OASIS-1 dataset ($C_{base}=32$, $m=4$). Mean and Standard deviation of Dice Coefficient (DC) and Average Symmetric Surface Distance (ASSD) are reported for different classes and the Average of all classes}
\resizebox{\columnwidth}{!}{
\begin{tabular}{lllllll}
\hline
SC  & DSL              & Cortex                                                       & SGM                                                          & WM                                                           & CSF                                                          & Mean                                                         \\ \hline
    &                  & \multicolumn{5}{c}{DC}                                                                                                                                                                                                                                                                                                   \\ \cline{3-7} 
Yes & Yes              & \begin{tabular}[c]{@{}l@{}}0.904\\±0.015\end{tabular} & \begin{tabular}[c]{@{}l@{}}0.941\\±0.014\end{tabular} & \begin{tabular}[c]{@{}l@{}}0.957\\±0.006\end{tabular} & \begin{tabular}[c]{@{}l@{}}0.929\\±0.032\end{tabular} & \begin{tabular}[c]{@{}l@{}}0.933\\±0.009\end{tabular} \\
Yes & No               & \begin{tabular}[c]{@{}l@{}}0.906\\±0.015\end{tabular} & \begin{tabular}[c]{@{}l@{}}0.939\\±0.016\end{tabular} & \begin{tabular}[c]{@{}l@{}}0.957\\±0.006\end{tabular} & \begin{tabular}[c]{@{}l@{}}0.919\\±0.036\end{tabular} & \begin{tabular}[c]{@{}l@{}}0.930\\±0.010\end{tabular} \\
No  & \textbackslash{} & \begin{tabular}[c]{@{}l@{}}0.878\\±0.016\end{tabular} & \begin{tabular}[c]{@{}l@{}}0.924\\±0.016\end{tabular} & \begin{tabular}[c]{@{}l@{}}0.947\\±0.006\end{tabular} & \begin{tabular}[c]{@{}l@{}}0.901\\±0.035\end{tabular} & \begin{tabular}[c]{@{}l@{}}0.912\\±0.010\end{tabular} \\ \hline
    &                  & \multicolumn{5}{c}{ASSD}                                                                                                                                                                                                                                                                                                 \\ \cline{3-7} 
Yes & Yes              & \begin{tabular}[c]{@{}l@{}}0.514\\±0.075\end{tabular} & \begin{tabular}[c]{@{}l@{}}0.516\\±0.170\end{tabular} & \begin{tabular}[c]{@{}l@{}}0.453\\±0.100\end{tabular} & \begin{tabular}[c]{@{}l@{}}0.396\\±0.131\end{tabular} & \begin{tabular}[c]{@{}l@{}}0.470\\±0.083\end{tabular} \\
Yes & No               & \begin{tabular}[c]{@{}l@{}}0.491\\±0.062\end{tabular} & \begin{tabular}[c]{@{}l@{}}0.520\\±0.188\end{tabular} & \begin{tabular}[c]{@{}l@{}}0.432\\±0.080\end{tabular} & \begin{tabular}[c]{@{}l@{}}0.442\\±0.131\end{tabular} & \begin{tabular}[c]{@{}l@{}}0.472\\±0.084\end{tabular} \\
No  & \textbackslash{} & \begin{tabular}[c]{@{}l@{}}0.675\\±0.079\end{tabular} & \begin{tabular}[c]{@{}l@{}}0.838\\±0.195\end{tabular} & \begin{tabular}[c]{@{}l@{}}0.538\\±0.081\end{tabular} & \begin{tabular}[c]{@{}l@{}}0.687\\±0.207\end{tabular} & \begin{tabular}[c]{@{}l@{}}0.685\\±0.100\end{tabular} \\ \hline
\end{tabular}}
\end{table}

\begin{table}[!h]
\caption{\label{tab2}Evaluation of model complexity ($C_{base}=32$, $m=4$) on OASIS-1 dataset. The mean value of number of parameters (NP), memory consumption (GM) and training time per step are reported}
\resizebox{\columnwidth}{!}{%
\tiny
\begin{tabular}{lllll}
\hline
SC  & DSL              & NP (m) & GM (GB) & TS (s) \\ \hline
%\multicolumn{5}{c}{Default Setting}                \\ \hline
%Yes & Yes              & 21.48  & 2.21    & 0.134  \\
%Yes & No               & 552.69 & 16.96   & 0.296  \\
%No  & \textbackslash{} & 21.47  & 2.17    & 0.111  \\ \hline
%\multicolumn{5}{c}{$C_{base}=32$, $m=4$}                  \\ \hline
Yes & Yes              & 11.84  & 2.41    & 0.126  \\
Yes & No               & 138.34 & 9.35    & 0.423  \\
No  & \textbackslash{} & 11.83  & 2.39    & 0.129  \\ \hline
\end{tabular}%
}
\end{table}

The mean and standard deviation of DC and ASSD are reported in Table \ref{tab1}, and the evaluations of model complexity are reported in Table \ref{tab2}. Firstly, it is seen from Table \ref{tab1} that compared to the model without SC, SC (with and without DSL) can increase the performance measured by both DC and ASSD significantly in all classes ($p<0.01$ measured by WSRT). Additionally, the use of DSL further increased the performance on the mean class results measured by DC significantly ($p<0.01$ measured by WSRT) and by ASSD non-significantly ($p=0.38$ measured by WSRT), compared to the model without DSL.

By achieving the best segmentation accuracy, the model with SC and DSL has very similar nP, GM and TS, compared to the model without SC and DSL as shown in Table \ref{tab2}. The model with SC but without DSL requires significantly larger NP, GM and longer TS, although has a similar performance with the model with both SC and DSL. Both Table \ref{tab1} and \ref{tab2} show the necessity of SC and DSL in achieving good segmentation performance with lower model complexity, less memory consumption and quicker training speed.  

\subsection{Comparison to the state-of-the-art methods}
Next, we compared the performance of our method ConvTransSeg (CTS) with the state-of-the-art methods. We chose one representative method with highly reported performance from each of the categories described in section \ref{sec:Related works}. These methods are FAT-Net \cite{wu2022fat} from the category \textit{Transformer in encoder}, SegTran \cite{li2021medical} from the category \textit{Transformer in decoder}, TransUnet \cite{chen2021transunet} from the category \textit{Transformer as a bridge}, and UTnet \cite{gao2021utnet} from the category \textit{Fused CNN and Transformer}, SwinUnet \cite{cao2021swin} from the category \textit{Pure Transformer}. Additionally, we also compared our model with the pure CNN method ResUNet \cite{zhang2018road}. For a fair comparison, data augmentation was not applied in the training process in any of the methods. Additionally, besides using the input size of $224\times224$, we also trained our method and ResUNet with input size of $256\times256$ for comparison with UTnet \cite{gao2021utnet}. The mean and standard deviation of DC and ASSD are reported in section \ref{sec:Segmentation accuracy} to compare the segmentation accuracy of all methods. Additionally, NP, GM and TS are reported in section \ref{sec:Model complexity} to compare the models’ complexity and training efficiency. Finally, some qualitative results are shown in section \ref{sec:Qualitative analysis} to demonstrate the advantages and drawbacks of each method visually. 

\subsubsection{Segmentation accuracy}
\label{sec:Segmentation accuracy}

% Please add the following required packages to your document preamble:
% \usepackage{graphicx}
\begin{table}[!h]
\caption{\label{tab3}Comparison of model performance on five public datasets. Mean and Standard deviation of Dice coefficient (DC) and Average Symmetric Surface Distance (ASSD) are reported. $\ast$ indicates that the method used $256\times256$ as the input size, otherwise $224\times224$ was used. Numbers in bold are the best methods in each input size. $\dag$ indicates that the result is significantly worse than the best results ($p<0.01$ measured by WSRT).}
\resizebox{\columnwidth}{!}{%
\begin{tabular}{llllll}
\hline
Model &
  ISIC &
  Bowl &
  Pannuke &
  CVC &
  OASIS \\ \hline
 &
  \multicolumn{5}{c}{DC} \\ \cline{2-6} 
ResUNet &
  \begin{tabular}[c]{@{}l@{}}0.831\\±0.195 $\dag$\end{tabular} &
  \begin{tabular}[c]{@{}l@{}}0.914\\±0.073\end{tabular} &
  \begin{tabular}[c]{@{}l@{}}0.512\\±0.190 $\dag$\end{tabular} &
  \begin{tabular}[c]{@{}l@{}}0.774\\±0.225 $\dag$\end{tabular} &
  \textbf{\begin{tabular}[c]{@{}l@{}}0.937\\±0.011\end{tabular}} \\
SwinUnet &
  \begin{tabular}[c]{@{}l@{}}0.877\\±0.145 $\dag$\end{tabular} &
  \begin{tabular}[c]{@{}l@{}}0.905\\±0.059 $\dag$\end{tabular} &
  \begin{tabular}[c]{@{}l@{}}0.516\\±0.186 $\dag$\end{tabular} &
  \begin{tabular}[c]{@{}l@{}}0.792\\±0.222 $\dag$\end{tabular} &
  \begin{tabular}[c]{@{}l@{}}0.933\\±0.010 $\dag$\end{tabular} \\
FAT-Net &
  \begin{tabular}[c]{@{}l@{}}0.884\\±0.139 $\dag$\end{tabular} &
  \begin{tabular}[c]{@{}l@{}}0.901\\±0.069 $\dag$\end{tabular} &
  \begin{tabular}[c]{@{}l@{}}0.495\\±0.197 $\dag$\end{tabular} &
  \begin{tabular}[c]{@{}l@{}}0.894\\±0.163\end{tabular} &
  \begin{tabular}[c]{@{}l@{}}0.775\\±0.008 $\dag$\end{tabular} \\
SegTran &
  \begin{tabular}[c]{@{}l@{}}0.874\\±0.161 $\dag$\end{tabular} &
  \begin{tabular}[c]{@{}l@{}}0.895\\±0.077 $\dag$\end{tabular} &
  \begin{tabular}[c]{@{}l@{}}0.526\\±0.188 $\dag$\end{tabular} &
  \begin{tabular}[c]{@{}l@{}}0.835\\±0.253 $\dag$\end{tabular} &
  \begin{tabular}[c]{@{}l@{}}0.906\\±0.010 $\dag$\end{tabular} \\
TransUnet &
  \begin{tabular}[c]{@{}l@{}}0.894\\±0.120\end{tabular} &
  \begin{tabular}[c]{@{}l@{}}0.900\\±0.063 $\dag$\end{tabular} &
  \begin{tabular}[c]{@{}l@{}}0.500\\±0.182 $\dag$\end{tabular} &
  \begin{tabular}[c]{@{}l@{}}0.872\\±0.173 $\dag$\end{tabular} &
  \begin{tabular}[c]{@{}l@{}}0.917\\±0.011 $\dag$\end{tabular} \\
CTS (ours) &
  \textbf{\begin{tabular}[c]{@{}l@{}}0.900\\±0.110\end{tabular}} &
  \textbf{\begin{tabular}[c]{@{}l@{}}0.917\\±0.065\end{tabular}} &
  \textbf{\begin{tabular}[c]{@{}l@{}}0.551\\±0.180\end{tabular}} &
  \textbf{\begin{tabular}[c]{@{}l@{}}0.900\\±0.137\end{tabular}} &
  \begin{tabular}[c]{@{}l@{}}0.933\\±0.010 $\dag$\end{tabular} \\
ResUNet$\ast$ &
  \begin{tabular}[c]{@{}l@{}}0.813\\±0.197 $\dag$\end{tabular} &
  \begin{tabular}[c]{@{}l@{}}0.919\\±0.063 $\dag$\end{tabular} &
  \begin{tabular}[c]{@{}l@{}}0.525\\±0.187 $\dag$\end{tabular} &
  \begin{tabular}[c]{@{}l@{}}0.781\\±0.227 $\dag$\end{tabular} &
  \textbf{\begin{tabular}[c]{@{}l@{}}0.943\\±0.011\end{tabular}} \\
UTnet$\ast$ &
  \begin{tabular}[c]{@{}l@{}}0.879\\±0.128 $\dag$\end{tabular} &
  \begin{tabular}[c]{@{}l@{}}0.837\\±0.243 $\dag$\end{tabular} &
  \begin{tabular}[c]{@{}l@{}}0.553\\±0.189 $\dag$\end{tabular} &
  \begin{tabular}[c]{@{}l@{}}0.862\\±0.184 $\dag$\end{tabular} &
  \begin{tabular}[c]{@{}l@{}}0.936\\±0.012 $\dag$\end{tabular} \\
CTS (ours)$\ast$ &
  \textbf{\begin{tabular}[c]{@{}l@{}}0.885\\±0.152\end{tabular}} &
  \textbf{\begin{tabular}[c]{@{}l@{}}0.925\\±0.066\end{tabular}} &
  \textbf{\begin{tabular}[c]{@{}l@{}}0.564\\±0.179\end{tabular}} &
  \textbf{\begin{tabular}[c]{@{}l@{}}0.889\\±0.158\end{tabular}} &
  \textbf{\begin{tabular}[c]{@{}l@{}}0.943\\±0.010\end{tabular}} \\ \hline
 &
  \multicolumn{5}{c}{ASSD} \\ \cline{2-6} 
ResUNet &
  \begin{tabular}[c]{@{}l@{}}9.432\\±20.595 $\dag$\end{tabular} &
  \begin{tabular}[c]{@{}l@{}}1.485\\±3.648 $\dag$\end{tabular} &
  \begin{tabular}[c]{@{}l@{}}25.422\\±25.227 $\dag$\end{tabular} &
  \begin{tabular}[c]{@{}l@{}}16.605\\±39.002 $\dag$\end{tabular} &
  \textbf{\begin{tabular}[c]{@{}l@{}}0.438\\±0.108\end{tabular}} \\
SwinUnet &
  \begin{tabular}[c]{@{}l@{}}5.279\\±6.862 $\dag$\end{tabular} &
  \begin{tabular}[c]{@{}l@{}}1.482\\±2.827 $\dag$\end{tabular} &
  \begin{tabular}[c]{@{}l@{}}27.921\\±28.299 $\dag$\end{tabular} &
  \begin{tabular}[c]{@{}l@{}}7.066\\±9.806 $\dag$\end{tabular} &
  \begin{tabular}[c]{@{}l@{}}0.454\\±0.095 $\dag$\end{tabular} \\
FAT-Net &
  \begin{tabular}[c]{@{}l@{}}5.153\\±6.742 $\dag$\end{tabular} &
  \begin{tabular}[c]{@{}l@{}}1.444\\±2.303 $\dag$\end{tabular} &
  \begin{tabular}[c]{@{}l@{}}23.567\\±25.987 $\dag$\end{tabular} &
  \begin{tabular}[c]{@{}l@{}}7.330\\±29.772 $\dag$\end{tabular} &
  \begin{tabular}[c]{@{}l@{}}0.933\\±0.095 $\dag$\end{tabular} \\
SegTran &
  \begin{tabular}[c]{@{}l@{}}6.736\\±18.198 $\dag$\end{tabular} &
  \begin{tabular}[c]{@{}l@{}}1.436\\±2.076 $\dag$\end{tabular} &
  \begin{tabular}[c]{@{}l@{}}24.392\\±24.193 $\dag$\end{tabular} &
  \begin{tabular}[c]{@{}l@{}}14.854\\±48.232 $\dag$\end{tabular} &
  \begin{tabular}[c]{@{}l@{}}0.725\\±0.132 $\dag$\end{tabular} \\
TransUnet &
  \begin{tabular}[c]{@{}l@{}}4.654\\±6.320 $\dag$\end{tabular} &
  \begin{tabular}[c]{@{}l@{}}1.406\\±1.750 $\dag$\end{tabular} &
  \begin{tabular}[c]{@{}l@{}}59.111\\±26.916 $\dag$\end{tabular} &
  \begin{tabular}[c]{@{}l@{}}5.886\\±12.744 $\dag$\end{tabular} &
  \begin{tabular}[c]{@{}l@{}}0.601\\±0.103 $\dag$\end{tabular} \\
CTS (ours) &
  \textbf{\begin{tabular}[c]{@{}l@{}}4.211\\±4.751\end{tabular}} &
  \textbf{\begin{tabular}[c]{@{}l@{}}1.125\\±1.642\end{tabular}} &
  \textbf{\begin{tabular}[c]{@{}l@{}}21.535\\±19.320\end{tabular}} &
  \textbf{\begin{tabular}[c]{@{}l@{}}4.003\\±6.878\end{tabular}} &
  \begin{tabular}[c]{@{}l@{}}0.469\\±0.095 $\dag$\end{tabular} \\
ResUNet$\ast$ &
  \begin{tabular}[c]{@{}l@{}}10.909\\±17.974 $\dag$\end{tabular} &
  \begin{tabular}[c]{@{}l@{}}1.475\\±2.620 $\dag$\end{tabular} &
  \begin{tabular}[c]{@{}l@{}}26.681\\±29.764 $\dag$\end{tabular} &
  \begin{tabular}[c]{@{}l@{}}16.643\\±39.871 $\dag$\end{tabular} &
  \begin{tabular}[c]{@{}l@{}}0.465\\±0.121 $\dag$\end{tabular} \\
UTnet$\ast$ &
  \begin{tabular}[c]{@{}l@{}}6.405\\±12.958 $\dag$\end{tabular} &
  \begin{tabular}[c]{@{}l@{}}5.083\\±26.011 $\dag$\end{tabular} &
  \begin{tabular}[c]{@{}l@{}}27.328\\±29.459 $\dag$\end{tabular} &
  \begin{tabular}[c]{@{}l@{}}8.567\\±26.379\end{tabular} &
  \begin{tabular}[c]{@{}l@{}}0.466\\±0.102 $\dag$\end{tabular} \\
CTS (ours)$\ast$ &
  \textbf{\begin{tabular}[c]{@{}l@{}}5.501\\±7.742\end{tabular}} &
  \textbf{\begin{tabular}[c]{@{}l@{}}1.167\\±1.695\end{tabular}} &
  \textbf{\begin{tabular}[c]{@{}l@{}}22.590\\±20.328\end{tabular}} &
  \textbf{\begin{tabular}[c]{@{}l@{}}5.202\\±9.038\end{tabular}} &
  \textbf{\begin{tabular}[c]{@{}l@{}}0.430\\±0.087\end{tabular}} \\ \hline
\end{tabular}%
}
\end{table}

From the results reported in Table \ref{tab3}, it is seen that our method achieved the best mean DC and mean ASSD on four out of five datasets (except for OASIS) using both input image sizes. Our method ranked the second after ResUNet and the third after ResUNet and SwinUnet on the OASIS dataset using DC and ASSD measurements respectively, when the input image size is $224 \times 224$. When a larger input image size ($256 \times 256$) is used, our method achieved the same performance as ResUnet and performed better than UT-net using DC measure, and it performed the best using ASSD measure. In conclusion, our method (CTS) is overall the best method when evaluated on different datasets with varying input image size based on DC and ASSD measures. 

We also ran statistical tests (WSRT) in comparing each method to the best method for each dataset indicated by $\dag$ in Table \ref{tab3}. It can be seen that our method is the best method in most datasets. Although it is not always significantly better ($p<0.01$ by WSRT) than the second best method, the second best method is not always the same in these datasets. 

\subsubsection{Model complexity}
\label{sec:Model complexity}

Table \ref{tab4} presents the measurements of NP, GM and TS for comparing the number of parameters, memory consumption and training speed respectively. It is seen that our method requires the least number of learning parameters compared to other Transformer based methods. Although the pure Transformer SwinUnet consumed the least GPU memory and had the quickest training speed, it performed worse than our method in 4 out of 5 datasets and similar performance on the remaining dataset (OASIS) in Table \ref{tab3}. More importantly, although our method is not the best method in terms of model complexity and training speed, pre-training is not needed in our method. The only Transformer based method that does not require pre-training is UTnet, but it has many more learnable parameters and performed worse in all datasets than our method. 

\begin{table}[!h]
\caption{\label{tab4}Comparison on model complexity on binary segmentation tasks. The mean value of number of parameters (NP), memory consumption (GM) and training time per step (TS) are reported}
\resizebox{\columnwidth}{!}{%
\tiny
\begin{tabular}{llll}
\hline
Model     & nP/m   & GM/GB & TS (s) \\ \hline
ResUNet   & 13.04  & 1.39  & 0.120  \\
SwinUnet  & 27.17  & 0.94  & 0.110  \\
FAT-Net   & 28.76  & 1.01  & 0.132  \\
SegTran   & 50.47  & 0.96  & 0.131  \\
TransUnet & 105.28 & 2.46  & 0.174  \\
CTS (ours)       & 21.48  & 2.21  & 0.139  \\
ResUNet$\ast$  & 13.04  & 1.39  & 0.161  \\
UTnet$\ast$    & 57.46  & 2.67  & 0.250  \\
CTS (ours)$\ast$      & 21.60  & 3.03  & 0.161  \\ \hline
\end{tabular}%
}
\end{table}

\subsubsection{Qualitative analysis}
\label{sec:Qualitative analysis}
To find the dataset on which the performances of different methods differ the largest, we calculated the sum of pair-wise absolute difference of the DC values of different methods for each dataset reported in Table \ref{tab3}. The CVC-ClinicDB dataset was identified using such a measure, hence some examples were extracted from this dataset for qualitative comparison. 

\begin{figure}[!h]
\centering
\includegraphics[width=0.85\linewidth]{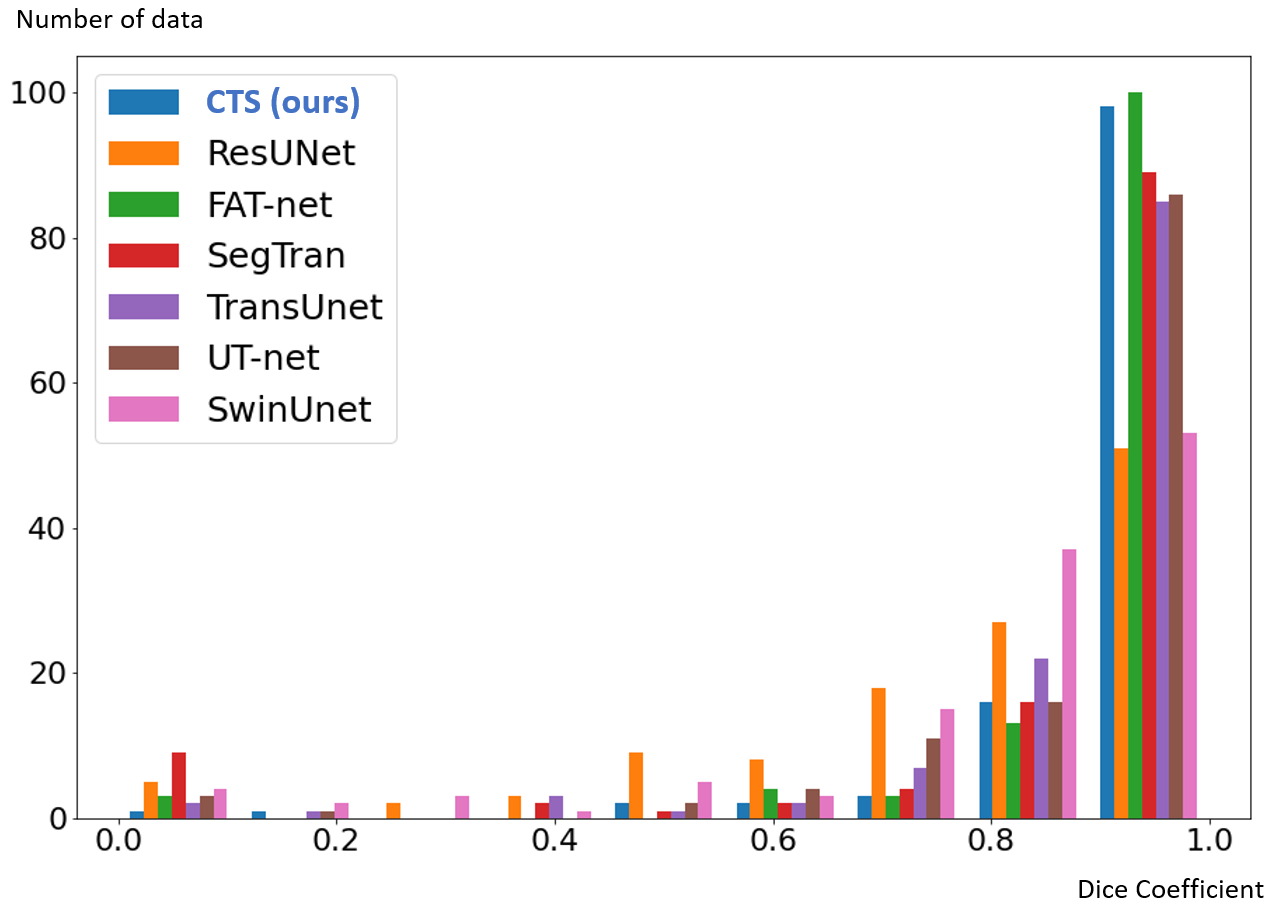}
\caption{\label{fig4}The distributions of DC values of different methods on CVC dataset}
\end{figure}

To visualize the performance differences between different methods, we firstly plot the distributions of DC values of all methods on CVC dataset in Fig \ref{fig4}. The plot illustrates that CTS (our method) and FAT-net have more high-quality segmentation results ($DC>0.9$) than other methods. ResUNet and SegTran have more low-quality segmentation results ($DC<0.1$) than other methods. SegTran also uses Transformer in decoder but different from our method, it reveals the strength of our proposed new model structure. 

From both Table \ref{tab3} and Fig \ref{fig4}, it can be seen that ResUNet performed much worse than all other Transformer-based methods on the CVC dataset. The CVC dataset is for polyp detection, which contains many challenging cases where the foreground and background share very similar image intensities and feature patterns. In this case, long range image dependencies and interactions play an important role in distinguishing foreground from background when they share similar features.

\begin{figure*}[!h]
\centering
\includegraphics[width=0.85\linewidth]{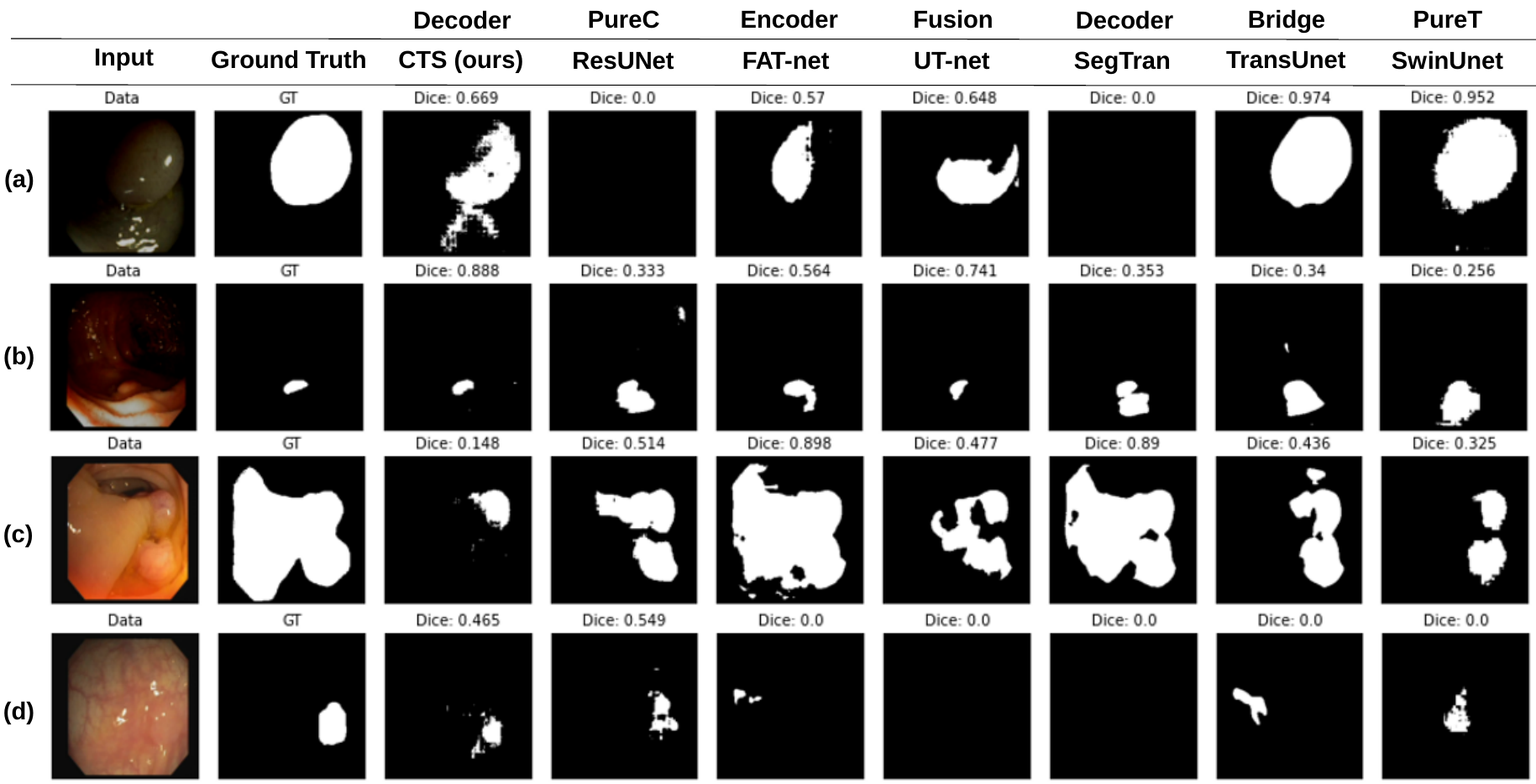}
\caption{\label{fig5}Some predictions on CVC-ClinicDB that show large difference between ResUNet and our method. The images from left to right is the original data, ground truth, predictions of our method, ResUNet, FAT-net, UT-net, SegTran, TransUnet, and SwinUnet. ‘Decoder’, ‘PureC’, ‘Encoder’, ‘Fusion’, ‘Bridge’, ‘PureT’ indicates the category \textit{Transformer in decoder}, \textit{Pure CNN}, \textit{Transformer in encoder}, \textit{Fused CNN and
Transformer}, \textit{Transformer as a bridge}, \textit{Pure Transformer} respectively}
\end{figure*}

Furthermore, we selected some representative cases where the DC values of different methods have large standard deviations, as shown in Fig \ref{fig5}. From Fig \ref{fig5} (a), it can be seen that when the background is dark, ResUNet produced no predictions. Our method was influenced by the noise (the reflection part), which resulted in over-segmentation in the bottom part. The results from both TransUnet and SwinUnet were more similar to the ground truth mask. However arguably, it is a challenging case even for human annotators to produce a reliable ground truth. In Fig \ref{fig5} (b), the target object shares very similar image intensity with a large part of the background. Our method produced the best results compared to other methods. Additionally, Fig \ref{fig5} (c) represents a case of large region of interest. In this case, SegTran and FAT-net performed well and our method under-segmented the target with the lowest DC value. However, our method managed to detect one of the poly regions correctly. The example in Fig \ref{fig5} (d) can be considered as the situation where the target is difficult to be distinguished from the background, where only ResUNet and our method can make some correct predictions in the target region. Although ResUNet has higher dice coefficient, our method captures the location of the target more correctly. Overall, SegTran (also using Transformer as decoder) performs the worst on most cases, which reveals the advantage of our new method of using Transformer as decoder.

\section{Discussion and conclusions}
In this work, we have proposed a hybrid model called ConvTransSeg that combines components of Transformer and CNN. By taking the advantages of both attention mechanism and CNN, our method can learn the long-term dependency of multi-resolution local features. We have evaluated our method and compared its segmentation accuracy, model complexity, and training efficiency with pure CNNs, pure Transformer, and other Transformer-CNN hybrid methods on five public datasets of different medical segmentation tasks. The results demonstrate that our method has the least number of learnable parameters with a quick convergence speed. More importantly, it is overall the best method for both DC and ASSD measures based on the results provided in section \ref{sec:Segmentation accuracy}. We also qualitatively assessed some representative segmentation results of all compared methods on the CVC-ClinicDB dataset. It demonstrates the superiority of our method in segmenting challenging images that the target and background are sharing similar intensities and feature patterns. However, our method is not robust enough to handle noisy regions and tends to predict more false positives on the noisy regions (e.g. Fig \ref{fig5} (a)).

In conclusion, different from other Transformer-based methods, by applying CNN as encoder and utilizing the encoded features in multi-resolution, our Transformer-based decoder can learn from the local representations in different image scales (i.e. both large patches and small patches). Such multi-resolution features can provide rich information for the Transformer decoder to make decisions, even when the dataset has small number of images. Hence, unlike other Transformer-based methods, our method does not require any pre-training to achieve similar or better performance, and the input size of our method is flexible. 

Currently, there are very few works on applying Transformer to capture 3D information, due to the limitation of GPU memory. Thus, in future work, we will improve the attention mechanism used in our method and extend it to 3D images without consuming too much computational resource. In addition, we will modify our method to improve its robustness to better handle noisy image regions.

\bibliographystyle{unsrt}
\bibliography{references}

\begin{thebibliography}{10}

\bibitem{shamshad2022transformers}
Fahad Shamshad, Salman Khan, Syed~Waqas Zamir, Muhammad~Haris Khan, Munawar
  Hayat, Fahad~Shahbaz Khan, and Huazhu Fu.
\newblock Transformers in medical imaging: A survey.
\newblock {\em arXiv preprint arXiv:2201.09873}, 2022.

\bibitem{kass1988snakes}
Michael Kass, Andrew Witkin, and Demetri Terzopoulos.
\newblock Snakes: Active contour models.
\newblock {\em International journal of computer vision}, 1(4):321--331, 1988.

\bibitem{gee1993elastically}
James~C Gee, Martin Reivich, and Ruzena Bajcsy.
\newblock Elastically deforming a three-dimensional atlas to match anatomical
  brain images.
\newblock 1993.

\bibitem{boykov2000interactive}
Yuri Boykov and Marie-Pierre Jolly.
\newblock Interactive organ segmentation using graph cuts.
\newblock In {\em International conference on medical image computing and
  computer-assisted intervention}, pages 276--286. Springer, 2000.

\bibitem{zhou2021review}
S~Kevin Zhou, Hayit Greenspan, Christos Davatzikos, James~S Duncan, Bram
  Van~Ginneken, Anant Madabhushi, Jerry~L Prince, Daniel Rueckert, and Ronald~M
  Summers.
\newblock A review of deep learning in medical imaging: Imaging traits,
  technology trends, case studies with progress highlights, and future
  promises.
\newblock {\em Proceedings of the IEEE}, 109(5):820--838, 2021.

\bibitem{ramachandran2019stand}
Prajit Ramachandran, Niki Parmar, Ashish Vaswani, Irwan Bello, Anselm Levskaya,
  and Jon Shlens.
\newblock Stand-alone self-attention in vision models.
\newblock {\em Advances in Neural Information Processing Systems}, 32, 2019.

\bibitem{ronneberger2015u}
Olaf Ronneberger, Philipp Fischer, and Thomas Brox.
\newblock U-net: Convolutional networks for biomedical image segmentation.
\newblock In {\em International Conference on Medical image computing and
  computer-assisted intervention}, pages 234--241. Springer, 2015.

\bibitem{long2015fully}
Jonathan Long, Evan Shelhamer, and Trevor Darrell.
\newblock Fully convolutional networks for semantic segmentation.
\newblock In {\em Proceedings of the IEEE conference on computer vision and
  pattern recognition}, pages 3431--3440, 2015.

\bibitem{zhang2018road}
Zhengxin Zhang, Qingjie Liu, and Yunhong Wang.
\newblock Road extraction by deep residual u-net.
\newblock {\em IEEE Geoscience and Remote Sensing Letters}, 15(5):749--753,
  2018.

\bibitem{he2016deep}
Kaiming He, Xiangyu Zhang, Shaoqing Ren, and Jian Sun.
\newblock Deep residual learning for image recognition.
\newblock In {\em Proceedings of the IEEE conference on computer vision and
  pattern recognition}, pages 770--778, 2016.

\bibitem{zhou2018unet++}
Zongwei Zhou, Md~Mahfuzur Rahman~Siddiquee, Nima Tajbakhsh, and Jianming Liang.
\newblock Unet++: A nested u-net architecture for medical image segmentation.
\newblock In {\em Deep learning in medical image analysis and multimodal
  learning for clinical decision support}, pages 3--11. Springer, 2018.

\bibitem{cciccek20163d}
{\"O}zg{\"u}n {\c{C}}i{\c{c}}ek, Ahmed Abdulkadir, Soeren~S Lienkamp, Thomas
  Brox, and Olaf Ronneberger.
\newblock 3d u-net: learning dense volumetric segmentation from sparse
  annotation.
\newblock In {\em International conference on medical image computing and
  computer-assisted intervention}, pages 424--432. Springer, 2016.

\bibitem{milletari2016v}
Fausto Milletari, Nassir Navab, and Seyed-Ahmad Ahmadi.
\newblock V-net: Fully convolutional neural networks for volumetric medical
  image segmentation.
\newblock In {\em 2016 fourth international conference on 3D vision (3DV)},
  pages 565--571. IEEE, 2016.

\bibitem{vaswani2017attention}
Ashish Vaswani, Noam Shazeer, Niki Parmar, Jakob Uszkoreit, Llion Jones,
  Aidan~N Gomez, {\L}ukasz Kaiser, and Illia Polosukhin.
\newblock Attention is all you need.
\newblock {\em Advances in neural information processing systems}, 30, 2017.

\bibitem{dosovitskiy2020image}
Alexey Dosovitskiy, Lucas Beyer, Alexander Kolesnikov, Dirk Weissenborn,
  Xiaohua Zhai, Thomas Unterthiner, Mostafa Dehghani, Matthias Minderer, Georg
  Heigold, Sylvain Gelly, et~al.
\newblock An image is worth 16x16 words: Transformers for image recognition at
  scale.
\newblock {\em arXiv preprint arXiv:2010.11929}, 2020.

\bibitem{chen2021transunet}
Jieneng Chen, Yongyi Lu, Qihang Yu, Xiangde Luo, Ehsan Adeli, Yan Wang, Le~Lu,
  Alan~L Yuille, and Yuyin Zhou.
\newblock Transunet: Transformers make strong encoders for medical image
  segmentation.
\newblock {\em arXiv preprint arXiv:2102.04306}, 2021.

\bibitem{valanarasu2021medical}
Jeya Maria~Jose Valanarasu, Poojan Oza, Ilker Hacihaliloglu, and Vishal~M
  Patel.
\newblock Medical transformer: Gated axial-attention for medical image
  segmentation.
\newblock In {\em International Conference on Medical Image Computing and
  Computer-Assisted Intervention}, pages 36--46. Springer, 2021.

\bibitem{wu2022fat}
Huisi Wu, Shihuai Chen, Guilian Chen, Wei Wang, Baiying Lei, and Zhenkun Wen.
\newblock Fat-net: Feature adaptive transformers for automated skin lesion
  segmentation.
\newblock {\em Medical Image Analysis}, 76:102327, 2022.

\bibitem{codella2019skin}
Noel Codella, Veronica Rotemberg, Philipp Tschandl, M~Emre Celebi, Stephen
  Dusza, David Gutman, Brian Helba, Aadi Kalloo, Konstantinos Liopyris, Michael
  Marchetti, et~al.
\newblock Skin lesion analysis toward melanoma detection 2018: A challenge
  hosted by the international skin imaging collaboration (isic).
\newblock {\em arXiv preprint arXiv:1902.03368}, 2019.

\bibitem{bernal2015wm}
Jorge Bernal, F~Javier S{\'a}nchez, Gloria Fern{\'a}ndez-Esparrach, Debora Gil,
  Cristina Rodr{\'\i}guez, and Fernando Vilari{\~n}o.
\newblock Wm-dova maps for accurate polyp highlighting in colonoscopy:
  Validation vs. saliency maps from physicians.
\newblock {\em Computerized medical imaging and graphics}, 43:99--111, 2015.

\bibitem{gamper2019pannuke}
Jevgenij Gamper, Navid Alemi~Koohbanani, Ksenija Benet, Ali Khuram, and Nasir
  Rajpoot.
\newblock Pannuke: an open pan-cancer histology dataset for nuclei instance
  segmentation and classification.
\newblock In {\em European congress on digital pathology}, pages 11--19.
  Springer, 2019.

\bibitem{marcus2007open}
Daniel~S Marcus, Tracy~H Wang, Jamie Parker, John~G Csernansky, John~C Morris,
  and Randy~L Buckner.
\newblock Open access series of imaging studies (oasis): cross-sectional mri
  data in young, middle aged, nondemented, and demented older adults.
\newblock {\em Journal of cognitive neuroscience}, 19(9):1498--1507, 2007.

\bibitem{chorowski2015attention}
Jan~K Chorowski, Dzmitry Bahdanau, Dmitriy Serdyuk, Kyunghyun Cho, and Yoshua
  Bengio.
\newblock Attention-based models for speech recognition.
\newblock {\em Advances in neural information processing systems}, 28, 2015.

\bibitem{devlin2018bert}
Jacob Devlin, Ming-Wei Chang, Kenton Lee, and Kristina Toutanova.
\newblock Bert: Pre-training of deep bidirectional transformers for language
  understanding.
\newblock {\em arXiv preprint arXiv:1810.04805}, 2018.

\bibitem{wu2016google}
Yonghui Wu, Mike Schuster, Zhifeng Chen, Quoc~V Le, Mohammad Norouzi, Wolfgang
  Macherey, Maxim Krikun, Yuan Cao, Qin Gao, Klaus Macherey, et~al.
\newblock Google's neural machine translation system: Bridging the gap between
  human and machine translation.
\newblock {\em arXiv preprint arXiv:1609.08144}, 2016.

\bibitem{hatamizadeh2022unetr}
Ali Hatamizadeh, Yucheng Tang, Vishwesh Nath, Dong Yang, Andriy Myronenko,
  Bennett Landman, Holger~R Roth, and Daguang Xu.
\newblock Unetr: Transformers for 3d medical image segmentation.
\newblock In {\em Proceedings of the IEEE/CVF Winter Conference on Applications
  of Computer Vision}, pages 574--584, 2022.

\bibitem{DBLP:conf/miccai/QiY0LWL019}
Kehan Qi, Hao Yang, Cheng Li, Zaiyi Liu, Meiyun Wang, Qiegen Liu, and Shanshan
  Wang.
\newblock X-net: Brain stroke lesion segmentation based on depthwise separable
  convolution and long-range dependencies.
\newblock In Dinggang Shen, Tianming Liu, Terry~M. Peters, Lawrence~H. Staib,
  Caroline Essert, Sean Zhou, Pew{-}Thian Yap, and Ali~R. Khan, editors, {\em
  Medical Image Computing and Computer Assisted Intervention - {MICCAI} 2019 -
  22nd International Conference, Shenzhen, China, October 13-17, 2019,
  Proceedings, Part {III}}, volume 11766 of {\em Lecture Notes in Computer
  Science}, pages 247--255. Springer, 2019.

\bibitem{shen2022automated}
Zhiqiang Shen, Hua Yang, Zhen Zhang, and Shaohua Zheng.
\newblock Automated kidney tumor segmentation with convolution and transformer
  network.
\newblock In {\em International Challenge on Kidney and Kidney Tumor
  Segmentation}, pages 1--12. Springer, 2022.

\bibitem{chen2021transattunet}
Bingzhi Chen, Yishu Liu, Zheng Zhang, Guangming Lu, and David Zhang.
\newblock Transattunet: Multi-level attention-guided u-net with transformer for
  medical image segmentation.
\newblock {\em arXiv preprint arXiv:2107.05274}, 2021.

\bibitem{ji2021multi}
Yuanfeng Ji, Ruimao Zhang, Huijie Wang, Zhen Li, Lingyun Wu, Shaoting Zhang,
  and Ping Luo.
\newblock Multi-compound transformer for accurate biomedical image
  segmentation.
\newblock In {\em International Conference on Medical Image Computing and
  Computer-Assisted Intervention}, pages 326--336. Springer, 2021.

\bibitem{zhang2021multi}
Yinglin Zhang, Risa Higashita, Huazhu Fu, Yanwu Xu, Yang Zhang, Haofeng Liu,
  Jian Zhang, and Jiang Liu.
\newblock A multi-branch hybrid transformer network for corneal endothelial
  cell segmentation.
\newblock In {\em International Conference on Medical Image Computing and
  Computer-Assisted Intervention}, pages 99--108. Springer, 2021.

\bibitem{oktay2018attention}
Ozan Oktay, Jo~Schlemper, Loic~Le Folgoc, Matthew Lee, Mattias Heinrich,
  Kazunari Misawa, Kensaku Mori, Steven McDonagh, Nils~Y Hammerla, Bernhard
  Kainz, et~al.
\newblock Attention u-net: Learning where to look for the pancreas.
\newblock {\em arXiv preprint arXiv:1804.03999}, 2018.

\bibitem{li2021medical}
Shaohua Li, Xiuchao Sui, Xiangde Luo, Xinxing Xu, Yong Liu, and Rick Goh.
\newblock Medical image segmentation using squeeze-and-expansion transformers.
\newblock {\em arXiv preprint arXiv:2105.09511}, 2021.

\bibitem{li2021more}
Yijiang Li, Wentian Cai, Ying Gao, and Xiping Hu.
\newblock More than encoder: Introducing transformer decoder to upsample.
\newblock {\em arXiv preprint arXiv:2106.10637}, 2021.

\bibitem{li2021gt}
Yunxiang Li, Shuai Wang, Jun Wang, Guodong Zeng, Wenjun Liu, Qianni Zhang, Qun
  Jin, and Yaqi Wang.
\newblock Gt u-net: A u-net like group transformer network for tooth root
  segmentation.
\newblock In {\em International Workshop on Machine Learning in Medical
  Imaging}, pages 386--395. Springer, 2021.

\bibitem{zhang2021transfuse}
Yundong Zhang, Huiye Liu, and Qiang Hu.
\newblock Transfuse: Fusing transformers and cnns for medical image
  segmentation.
\newblock In {\em International Conference on Medical Image Computing and
  Computer-Assisted Intervention}, pages 14--24. Springer, 2021.

\bibitem{gao2021utnet}
Yunhe Gao, Mu~Zhou, and Dimitris~N Metaxas.
\newblock Utnet: a hybrid transformer architecture for medical image
  segmentation.
\newblock In {\em International Conference on Medical Image Computing and
  Computer-Assisted Intervention}, pages 61--71. Springer, 2021.

\bibitem{karimi2021convolution}
Davood Karimi, Serge~Didenko Vasylechko, and Ali Gholipour.
\newblock Convolution-free medical image segmentation using transformers.
\newblock In {\em International Conference on Medical Image Computing and
  Computer-Assisted Intervention}, pages 78--88. Springer, 2021.

\bibitem{cao2021swin}
Hu~Cao, Yueyue Wang, Joy Chen, Dongsheng Jiang, Xiaopeng Zhang, Qi~Tian, and
  Manning Wang.
\newblock Swin-unet: Unet-like pure transformer for medical image segmentation.
\newblock {\em arXiv preprint arXiv:2105.05537}, 2021.

\bibitem{liu2021swin}
Ze~Liu, Yutong Lin, Yue Cao, Han Hu, Yixuan Wei, Zheng Zhang, Stephen Lin, and
  Baining Guo.
\newblock Swin transformer: Hierarchical vision transformer using shifted
  windows.
\newblock In {\em Proceedings of the IEEE/CVF International Conference on
  Computer Vision}, pages 10012--10022, 2021.

\bibitem{huang2021missformer}
Xiaohong Huang, Zhifang Deng, Dandan Li, and Xueguang Yuan.
\newblock Missformer: An effective medical image segmentation transformer.
\newblock {\em arXiv preprint arXiv:2109.07162}, 2021.

\bibitem{yan2022after}
Xiangyi Yan, Hao Tang, Shanlin Sun, Haoyu Ma, Deying Kong, and Xiaohui Xie.
\newblock After-unet: Axial fusion transformer unet for medical image
  segmentation.
\newblock In {\em Proceedings of the IEEE/CVF Winter Conference on Applications
  of Computer Vision}, pages 3971--3981, 2022.

\end{thebibliography}

\end{document}